\documentclass[lettersize,journal]{IEEEtran}
\usepackage{amsmath,amsfonts}
\usepackage{algorithmic}
\usepackage{algorithm}
\usepackage{array}
\usepackage[caption=false,font=normalsize,labelfont=sf,textfont=sf]{subfig}
\usepackage{textcomp}
\usepackage{stfloats}
\usepackage{url}
\usepackage{verbatim}
\usepackage{graphicx}
\usepackage{cite}
\usepackage{booktabs}
\usepackage{multirow}
\usepackage{colortbl}  
\usepackage{makecell} 
\usepackage[table]{xcolor} 
\usepackage{afterpage}
\usepackage{lipsum} 
\usepackage{mathrsfs}

\begin{document}

\title{MuseumMaker: Continual Style Customization without Catastrophic Forgetting}

\author{Chenxi Liu$^{\dagger}$, Gan Sun$^{*}$~\IEEEmembership{Member,~IEEE}, Wenqi Liang$^{\dagger}$, Jiahua Dong, Can Qin, Yang Cong

\thanks{Chenxi Liu and Wenqi Liang are with the State Key Laboratory of Robotics, the Institutes for Robotics and Intelligent Manufacturing, Chinese Academy of Sciences, Shenyang 110169, China, and also
with the University of Chinese Academy of Sciences, Beijing 100049, China. (liuchenxi0101, liangwenqi0123@gmail.com.)}
\thanks{Gan Sun and Yang Cong are with the School of Automation Science and Engineering, South China University of Technology, Guangzhou, 510640, China. }
\thanks{Jiahua Dong is with the Mohamed bin Zayed University of Artificial Intelligence, Abu Dhabi, United Arab Emirates. (dongjiahua1995@gmail.com.)}
\thanks{Can Qin is with the Salesforce AI Research, Palo Alto, CA, 94301, USA.}

\thanks{$^{\dagger}$These authors contributed equally to this work.}
\thanks{$^{*}$The corresponding author is \emph{Prof. Gan Sun}.}}

\markboth{Journal of \LaTeX\ Class Files,~Vol.~14, No.~8, August~2021}%
{Shell \MakeLowercase{\textit{et al.}}: A Sample Article Using IEEEtran.cls for IEEE Journals}


\maketitle

\begin{abstract}
Pre-trained large text-to-image (T2I) models with an appropriate text prompt has attracted growing interests in customized images generation field. However, catastrophic forgetting issue make it hard to continually synthesize new user-provided styles while retaining the satisfying results amongst learned styles. In this paper, we propose \emph{MuseumMaker}, a method that enables the synthesis of images by following a set of customized styles in a never-end manner, and gradually accumulate these creative artistic works as a \emph{Museum}. When facing with a new customization style, we develop a style distillation loss module to extract and learn the styles of the training data for new image generation. It can minimize the learning biases caused by content of new training images, and address the catastrophic overfitting issue induced by few-shot images. To deal with catastrophic forgetting amongst past learned styles, we devise a dual regularization for shared-LoRA module to optimize the direction of model update, which could regularize the diffusion model from both weight and feature aspects, respectively. Meanwhile, to further preserve historical knowledge from past styles and address the limited representability of LoRA, we consider a task-wise token learning module where a unique token embedding is learned to denote a new style. As any new user-provided style come, our \emph{MuseumMaker} can capture the nuances of the new styles while maintaining the details of learned styles. Experimental results on diverse style datasets validate the effectiveness of our proposed \emph{MuseumMaker} method, showcasing its robustness and versatility across various scenarios.              



\end{abstract}


\begin{figure}[t]
	\centering
	\includegraphics[width=3.5in]
	{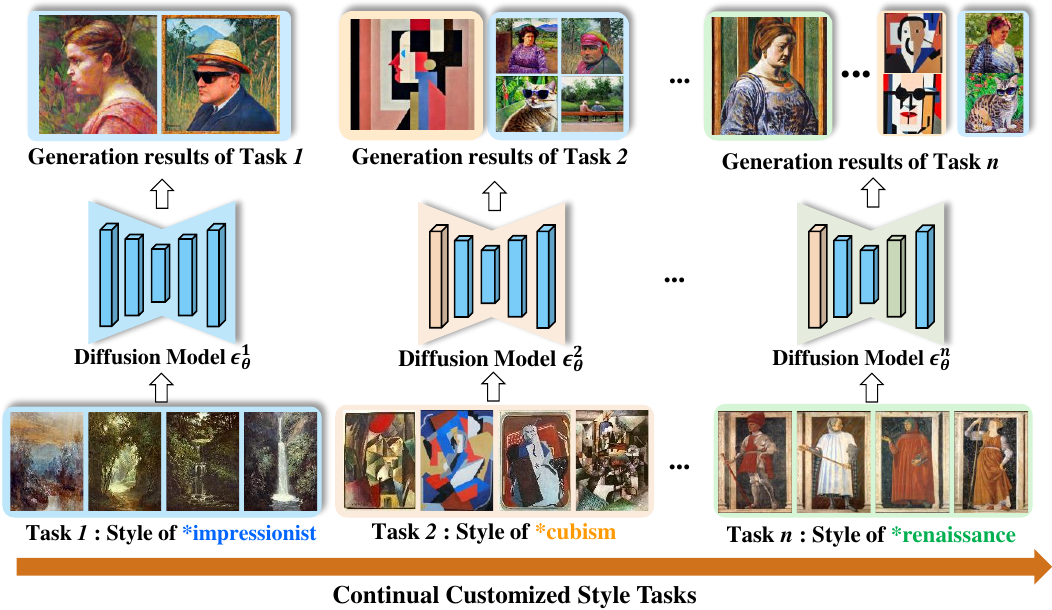}
 	\vspace{-20pt}
	\caption{Motivation of our proposed MuseumMaker model. The desired style (such as impressionist or cubism)  for each generated text-to-image task can be customized by the user with a few images. Our MuseumMaker model can continually incorporate the diverse styles without catastrophic forgetting, and further accumulate these creative works as a private Museum.   
 } 
	\label{fig: introduction}
	\vspace{-10pt}
\end{figure}
\section{Introduction}
\IEEEPARstart{T}{ext-to-image} (T2I) models have emerged in recent studies. Generative models based on diffusion model\cite{b2}\cite{b36}\cite{b68} have demonstrated remarkable efficacy and flexibility in the field of text-to-image generation. Amongst these methods, Stable Diffusion~\cite{b2} stands out for its ability to generate high quality images through simple language descriptions, which advances the application of generative models to new heights. Based on Stable Diffusion, the area of image generation has attracted widespread attentions, which also promotes the research of following areas, such as image super-resolution\cite{b62}\cite{b63} and image restoration\cite{b64}\cite{b65}.

Although most recent models\cite{b66}\cite{b67} focus on boosting their generative capabilities with natural languages, these models fail to control the exact texture and color palettes of generation images, when adding popular or customized styles (\emph{e.g.}, ``impressionist'') to the input text prompt. A naive way is to fine-tune the T2I diffusion model using few images of the given ``impressionist'' style, whereas this manner cannot synthesize various images of different specific styles in a never-ending manner. For example, as shown in Fig.~\ref{fig: introduction}, when the user inputs a prompt such as \emph{``a cat wearing sunglasses in the style of impressionism"} to generate an impressionist image of the cat, a naive way is to fine-tune the diffusion model with images in impressionist style. Subsequently, if the user tries to augment this diffusion model with a new style, \emph{e.g.}, ``cubism'', a simple method is to fine-tune the diffusion model with both impressionistic and cubist images. As users continuously seek to incorporate new styles, the large computational requirements and ever-increasing training times swiftly become impractical. Moreover, there is a high probability for diffusion model to overfit to the new user-provided styles after continual fine-tuning operator, resulting in unsatisfactory artistic generation performance\cite{hu2023phasic}.   



Inspired by the aforementioned practical scenarios, we assume that the pre-trained large T2I diffusion model receives data of different customized styles in a streaming manner, as shown in Fig.~\ref{fig: introduction}. In the scenario of continuous style stream, we aim to establish a continual style customization method with T2I diffusion model, and maintain original generative capacity even after extensive fine-tuning. To achieve this, the T2I diffusion models need to extract pure style features rather than intricate image content, and solve the following key challenges:
\begin{itemize}

\item ``\textbf{Catastrophic Overfitting}'', \emph{i.e.}, the ability to learn pure features stylistic representation from the images provided from users, instead of influencing by the complex and intricate content of images. When learning a new style from a limited set of images, how to overcome the problem of overfitting to the specific contents of user-provided images is a common thought of style learning. For instance, if the diffusion model has learned an ``impressionist'' style, which contains a significant number of pictures with mountains in training data. When we input a prompt like, \emph{``a smiling man in impressionism''}, the model may neglect or remove the concept of \textit{``man''} and instead output a picture of mountains in impressionism. 

\item ``\textbf{Catastrophic Forgetting}'', \emph{i.e.}, the knowledge of various styles obtained by diffusion model tends to be forgotten when incorporating with new styles. For this aspect, the personalized style knowledge forgetting learned from prior styles should be concerned. Boosting the ability to continually generate images of various styles without revisiting old data is a crucial consideration in the realm of continual style customization. For instance, the user is able to generate images of ``impressionism'' without accessing the training data, even after long periods of fine-tuning for new styles.

\end{itemize}



Inspired by the aforementioned considerations, we present a continual style learning approach based on T2I diffusion model in this work. Our proposed method aims to enable the continuous learning of new styles for generation purposes without accessing past datasets, which is designed to reduce memory consumption and compute efficiently. To address the two aforementioned challenges, we develop a continual style customization for diffusion model \emph{i.e}, MuseumMaker, which could continuously adapt upcoming new styles and purely learn stylistic features from user-provided data. To tackle the catastrophic overfitting problem, we introduce a \underline{S}tyle \underline{D}istillation \underline{L}oss (SDL) module, which decouples the style and content of images by extracting the representation of styles from the whole dataset. The SDL module enables the diffusion model to concentrate on acquiring the stylistic attributes of images while reducing the influence of specific content of images. To mitigate catastrophic forgetting, we propose a \underline{D}ual \underline{R}regularization for shared-\underline{LoRA} (DR-LoRA) module, which is designed to facilitate the smooth transfer of knowledge from old styles. This module incorporates LoRA weight regularization and stylistic feature regularization to preserve knowledge acquired from previous styles. All the customization tasks share a single set of LoRA parameters with DR-LoRA module to minimize memory consumption. Additionally, to address the limitation posed by the limited parameters of LoRA, a \underline{T}ask-wise \underline{T}oken \underline{L}earning (TTL) module is develops to learn a distinct token for each style, which enables generation of different styles. The tokens corresponding to past styles are stored to further mitigate the catastrophic forgetting problem. Finally, we demonstrate the validation of our MuseumMaker through extensive experiments, and conduct ablation studies to emphasize the contribution of each module in our MuseumMaker.

To summarise, the main contributions of this paper are as follows:
\begin{itemize}
\item We take an earlier attempt to propose continual style customization with pre-trained large T2I diffusion model, \emph{i.e., MuseumMaker}, which enables continual learning of various styles and effectively mitigates the catastrophic forgetting issue amongst past styles. 
\item To deal with the issue of catastrophic overfitting, we introduce a style distillation loss module to distill the style representation from the entire dataset into the latent representation generated from each image, which could overcome the problem of learning bias to the content of the images.
\item  To address the catastrophic forgetting issue, we devise a dual regularization for shared-LoRA module and a task-wise token learning module, which could maintain the style knowledge from previous learned styles. Extensive experiments have confirmed the significant performance improvements and effectiveness of our proposed {MuseumMaker}.
\end{itemize}

We organize the rest of the paper as follows: the first section briefly introduces some related work. The second section revisits the text-to-image diffusion model and defines the problem of continual style learning for diffusion model. Then, we describe our methods in detail in third section. To the end, we conduct various experiments to evaluate our proposed method, followed by the conclusion.

\section{Related Work}
\subsection{Continual Learning}
The field of continual learning has attracted significant attention in recent years. Early approaches to continual learning focus on regularization-based methods, such as EWC\cite{b11} and SI\cite{b12}. These methods introduce additional regularization terms to penalize the changes of parameters, which are important for preserving previously acquired knowledge. Subsequent works have explored extensions and variations of these techniques, including strategies for better approximating parameter importance \cite{b13}\cite{b15} and methods for considering the heterogeneous forgetting across different classes\cite{dong2023iccv} \cite{nooneleft}.

Another notable direction is experience replay, where a subset of past data is stored and replayed during training on new tasks to mitigate forgetting \cite{b16}. Variations of this approach include strategies for constructing and exploiting the memory buffer\cite{b17}\cite{b29}, and leveraging generated data from generative models to augment or replace the replay buffer\cite{b18}\cite{b19}.
Optimization-based approaches have also been explored, such as gradient projection methods\cite{congtcsvt}\cite{b30} and meta-learning strategies\cite{b21}\cite{b22}. These techniques aim to directly manipulate the optimization process or learn inductive biases that facilitate continual learning.
Architectural innovations also play a role in continual learning, with methods such as parameter isolation\cite{b24}\cite{b25} , dynamic architectures\cite{b26}\cite{b27}, and modular networks\cite{b31}.
More recently, there has been a growing interest in representation-based approaches that leverage category-guided learning\cite{dong2023inor} and large-scale pre-training\cite{b33}\cite{b34} to obtain robust and transferable representation for continual learning. These methods seek to exploit the inherent advantages of self-supervised and pre-trained representation, such as improved generalization and robustness to catastrophic forgetting.

Despite significant progress, continual learning remains an active area of research. In this area, continual style learning for diffusion model is still a matter of concern, where various stylistic feature representation are hard to merge into a special diffusion model. Continual generation for different styles of images remains a challenge for diffusion model.

\subsection{Text to Image Generation}
The emergence of diffusion models has ushered in a new era of text-to-image (T2I) generation, with state-of-the-art methods achieving unprecedented levels of photorealism and caption-image alignment. GLIDE\cite{b35} pioneers the application of diffusion models to this task, which adopts classifier-free guidance by replacing class labels with text prompts. Imagen \cite{b36} further improves upon this approach by leveraging pre-trained language models as text encoders, enabling the use of rich textual representation learned from large-scale corpora.

Other advanced works explore diffusion models in latent space, rather than operating directly on pixel space. Stable Diffusion\cite{b2} is trained with a large amount of data based on latent diffusion model (LDM), which employs a VQ-GAN\cite{b37} for latent representation and adds text as conditional information to the denoising process. DALL-E 2\cite{b38} inverts the CLIP\cite{b39} image encoder with a diffusion model, learning a prior to connect the text and the features of images in the latent space. This text-image latent prior is found to be crucial for performance.

Building upon these pioneering efforts, subsequent studies have sought to enhance T2I diffusion models in various ways. These improvements encompass model architectures\cite{b40,b41,b42}, spatial control via sketches or spatio-textual representation\cite{b43}\cite{b44}\cite{qin2023unicontrol}, textual inversion for controlling novel concepts\cite{b3}, and retrieval mechanisms\cite{b46,b47,b48} to handle out-of-distribution scenarios. To further tackle the issue of continual learning for diffusion, \cite{b60}\cite{create} achieve the continuous generation of personalized concepts. However, the methods for continuous style learning and image generation are still lack of research.

\subsection{Image Style Learning}
Capturing expressive representation of image styles is pivotal for achieving high-fidelity style learning. Early techniques like texture synthesis\cite{b49}\cite{b50} and non-photorealistic rendering (NPR)\cite{b51}\cite{b52} explore artistic expression through computational methods. However, these methods still face a constraint in transfer quality, universality, and feature extraction capabilities.



Generative Adversarial Networks (GANs) have emerged as a transformative paradigm for style representation learning. \cite{b57} proposes the DCGAN model to integrate a convolutional neural network into the original GAN framework, and adopts an encoder-decoder structure to bolster stability and generation quality. \cite{b58} introduces Wasserstein GAN (WGAN), leveraging the Earth Mover (EM) distance as a loss function to measure distributional discrepancies. With the recent rapid advancement of diffusion models, extensive methods for image style learning based on diffusion models have emerged. \cite{b59} treats the style information of images as trainable textual descriptions for model learning. \cite{b10} fine-tunes the U-Net architecture to enable the model to learn the stylistic information from images, and generates images in specific styles. However, these exciting style learning methods are unable to tackle a never-ending streaming manner of styles, which constrains the further application of diffusion model to generate images of various styles.

\section{Preliminaries}
\subsection{Revisiting Text-to-Image Diffusion Model}\label{chap:3.1} 
The diffusion model \cite{ddpm} represents a probabilistic approach utilized for image generation, which generates a image by gradually denoising a noise graph from a Gaussian distribution. As exemplified by the Stable Diffusion (SD) technique \cite{b2}, the primary objective of the diffusion model is to learn a trajectory to generate the well patterned and distributed data from a random noise graph.  Specifically, SD relies on a pre-trained text encoder ${\psi}$ from CLIP\cite{b39}, a latent encoder $\mathcal{F}$, a decoder $\mathcal{G}$ and a denoising U-Net $\epsilon_\theta$. Given a text prompt $p$, a timestep $t \in \text{Uniform}(1,T)$ and a random Gaussian noise graph $\varepsilon \in \mathcal{N}(0,\mathbf{I})$, the text embedding can be computed as $c = \psi(p)$ ,and subsequently, the image $\hat{x} $ can be generated by $\hat{x} = \mathcal{G}(\epsilon_\theta(c,t))$. Formally, the training process of SD can be defined as follow:

\begin{align}
        \label{eq: eq12}
       \mathcal{L}_{SD}=\mathbb{E}_{z,c,\varepsilon,t}[\|\varepsilon-\epsilon_\theta(z_t|c,t)\|_2^2],
\end{align}
where $z_t$ is the image latents encoded by $\mathcal{E}$ at timestep $t$, and $\theta$ represents the parameters of denoising model.

\subsection{Problem Definition}\label{chap:3.2} 


Suppose that a series of continuous tasks for Text-to-Image (T2I) diffusion model are denoted as $\mathbf{D}^K=\{\mathcal{D}^{1}_{n_1},\mathcal{D}^{2}_{n_2},\cdots,\mathcal{D}^{K}_{n_K}\}$, where $K$ represents the number of consecutive tasks and $n_k$ denotes the number of samples per task. Consequently, we can define all the generation tasks as  $\mathcal{T} = \{\mathcal{T}^k\}_{k=1}^{K}$. For the $k$-th learning task, $\mathcal{T}^k$ corresponds to a dataset $\mathcal{D}^{k}_{n_k} = \{x^k_i,p_i^k\}^{n_k}_{i=1}$, where $x^k_i \in \mathcal{X}^k$ and $p_i^k \in \mathcal{P}^k$ denote the $i$-th image and its corresponding prompt. Different from existing works\cite{b10}\cite{b9}, we consider a scenario that users provide consecutive images of different specific styles for T2I diffusion model, namely task $\mathcal{T}^k$. For each task, the number of pairs of image $x^k_i$ and prompt $p^k_i$ with $n_k$ are imposed into the continual style customization diffusion model, MuseumMaker. Considering the problem of privacy and memory limitation, the dataset of past tasks is unavailable. In this scenario, the model we proposed is able to incorporate the new style while ensuring the memory of past encountered styles. In details, the problem of continual style customization can be formulated as follows:
\begin{align}
        \label{eq: eq1} 
        \mathcal{L}^k_{\mathrm{SD}} =\frac1{\mathcal{B}}\sum_{i=1}^\mathcal{B} &\mathbb{E}_{z_i^k,c_i^k,\varepsilon,t}\left[\|\varepsilon-\epsilon_\theta^k(z_{i,t}^k|c_i^k)\|_2^2\right],\\
        \mathcal{L}_{\mathrm{CSA}} &=\sum_{k=1}^K(\mathcal{L}^k_{\mathrm{SD}}+\lambda\mathcal{L}_{\mathrm{c}}^k),
\end{align}
where $ \epsilon $ denotes the denoising U-Net, $ \theta $ represents its trainable parameters, and $\mathcal{B}$ is the batch size during train process. $(z_{i,t}^{k},c_i^k)$ respectively represent the corresponding conditional image latents at timestep $t$ and text embeddings on the $k$-th task, where $c_i^k = \psi(p_i^k)$. $\mathcal{L}^k_{c}$ is a loss proposed under the constraint of past prior knowledge, aiming at facilitating continuous learning of the model. Further explanation on $\mathcal{L}^k_{c}$ will be provided in the subsequent sections.
\begin{figure*}[t]
	\centering
	\includegraphics[width=7.1in]
    {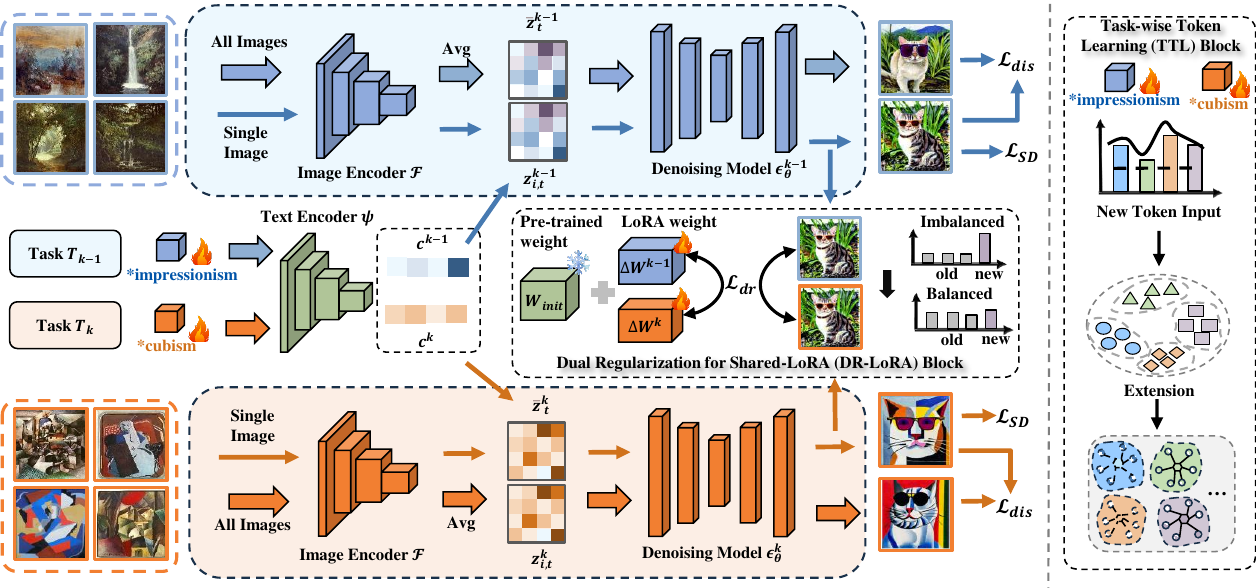}
 	\vspace{-10pt}
	\caption{Illustration of our proposed continual style customization for T2I diffusion model, \emph{i.e.,} MuseumMaker. It mainly contains a Dual Regularization for Shared-LoRA (DR-LoRA) module to regularize the optimization of model from both weight and feature aspects, a Task-wise Token Learning (TTL) module to store the text embedding of each style learning to reduce forgetting and a Style Distillation Loss module (SDL) to make the model focus on the style of learning images.} 
	\label{fig: framework}
	\vspace{-10pt}
\end{figure*}

\section{Our Proposed Method}
In this section, we firstly present the overview of our proposed MuseumMaker in Sec.~\ref{chap:4.1}, which is illustrated in Fig.~\ref{fig: framework}. In Sec.~\ref{chap:4.4}, we fristly present a style distillation loss to overcome the problem of catastrophic overfitting when encountering with a new style customization task. To tackle the catastrophic forgetting issue, we introduce a dual regularization for shared-LoRA module in Sec.~\ref{chap:4.2}, and a task-wise token learning module in Sec.~\ref{chap:4.3}. 

\subsection{Overview}\label{chap:4.1} 
As illustrated in Fig. \ref{fig: framework}, we develop a progressive continual style customization method with T2I diffusion model (\emph{i.e.,} MuseumMaker), which aims to retain knowledge of past learned styles while incorporating with new customized styles. To address the issue of catastrophic overfitting to image content, we devise a \underline{S}tyle \underline{D}istillation \underline{L}oss (SDL) module for each new style, which distills the mean latent features across all images with individual image latent features to reduce the influence of image content during training. The direction of optimization is corrected to style of images by SDL, ensuring that the model focuses on style of images and alleviates overfitting to content. Considering the issue of catastrophic forgetting in continual style learning, we develop a \underline{D}ual \underline{R}regularization for shared-\underline{LoRA} (DR-LoRA) module and a \underline{T}ask-wise \underline{T}oken \underline{L}earning (TTL) module. These two modules intend to transfer the knowledge from old model to current model and attain a unique token embedding for each specific style. Overall, MuseumMaker could offer a comprehensive solution for continual style customization.



\subsection{Style Distillation Loss}\label{chap:4.4} 
Given a reference style customization task with $8\sim10$ images, continual customized style learning aims to extract its artistic style information while explicitly removing the content information, which also emerges as a key area of focus in image style transfer~\cite{stylediffusion}\cite{disentanglement}. One of feasible solutions is to explicitly defines the high-level features as content and the feature correlations (\emph{i.e.}, Gram matrix) as style \cite{stylediffusion}. However, this method relies on additional networks, which is not feasible enough to deploy to diffusion model and not available for continual style customization. Therefore, we introduce an easily deployable style distillation loss module, which addresses catastrophic overfitting and captures the pure style representations to prevent the disruption of original concepts in diffusion model.


To capture pure style representation from user-provided images and guide the model towards learning style features, we propose a straightforward yet highly implementable method in the continual style customization setting, \emph{i.e.}, \underline{S}tyle \underline{D}istillation \underline{L}oss (SDL) module. Specifically, given a set of input images $\begin{aligned}\mathcal{X}^k=\{x^k_{1},x^k_{2},...x^k_{n_k}\}\end{aligned}$ on the $k$-th style customization task. We encode all images using an VAE \cite{vae} image encoder $\mathcal{F}$ to obtain their feature representation ${\mathcal{Z}^k}=\{z_{i}^k=\mathcal{F}(x_i^k)|x_i^k\in{\mathcal{X}}^k\}$, and then compute the mean of all feature representation as the representation of the image style of specific dataset. The latent features of images in the $k$-th task can be represented as follows:
\begin{align}
        \label{eq: eq8}
        \bar{z}_t^{k}=\frac1{n_k}\sum_{i=1}^{n_k}\mathcal{F}(x_{i}^k).
\end{align}
By utilizing both the mean feature and individual image features to train unique styles, we then obtain a distillation loss during the denoising process:
\begin{align}
        \label{eq:eq9}
        \mathcal{L}_{\mathrm{SDL}} = \frac1{{\mathcal{B}}}\sum_{i=1}^\mathcal{B}\rho(\mathcal{F}(x_{i}^k)/\tau)\mathrm{log}(\frac{\rho(\mathcal{F}(x_{i}^k)/\tau)}{\rho(\bar{z}_t^{k}/\tau)}),
\end{align}
where $\tau$ is a temperature hyperparameter, and $\rho$ represents the softmax function. Considering the Kullback-Leibler (KL) divergence loss between the mean latent features of all images and the latent features of individual images, SDL demonstrates its efficacy in reducing the probability of model overfitting to specific image content while learning image styles. This mechanism ensures the capacity of model to effectively concentrate on the diverse styles inherent in images.


\subsection{DR-LoRA}\label{chap:4.3} 
To rapid learn with extracted user-provided style from a small number of images, one of the popular strategies is fine-tuning T2I diffusion model with low-rank adaptation(LoRA)~\cite{b8}, which has demonstrated its effectiveness in various text-to-image diffusion models~\cite{b7}\cite{b60}. Inspired by LoRA, we adopt a shared-LoRA strategy for continual customized styles learning in this work, and intend to significantly reduce the computational consumption usage. However, a wide and disparate distribution existing in style data stream poses a formidable hurdle in the continual style customization setting, when the diffusion model tries to preserve the style knowledge among pre-trained model. To tackle this issue, we devise a \underline{D}ual \underline{R}regularization for this \underline{LoRA} module (DR-LoRA) to the balance the learning of different styles, which simultaneously considers the knowledge transfer of style distribution from both weight manifold and feature representation manifold. 

Specifically, LoRA achieves fine-tuning of pre-trained large T2I models by freezing original weight while learning a pair of rank-decomposition matrices. Formally, this can be expressed as: $\mathbf{W}' = \mathbf{W} + \Delta{\mathbf{W}}$, where $\Delta{\mathbf{W}} = \mathbf{AB}^T$. LoRA effectively reduces the number of parameters required during training by fine-tuning matrices $\mathbf{A}$ and $\mathbf{B}$, and makes a swift adaptation to new styles. However, without constraining the optimization of diffusion model, the model above tends to learn incoming style data and further causes the catastrophic forgetting. Therefore, we consider the weight regularization for LoRA weight, in which we compute the offset of LoRA between adjacent tasks as a loss function to optimize the direction of model updating. Mathematically, we denote the weight manifold regularization of the $k$-th task as follows:
\begin{equation}
\resizebox{0.47\textwidth}{!}{%
    $\displaystyle
    \begin{aligned}
        \label{eq: lp}
        \mathcal{L}_{w} &= \frac{1}{L} \sum_{l=1}^{L} \left[1 - \mathrm{Sim}\left(\text{Flatten}(\Delta{\mathbf{W}_l^{k-1}}), \text{Flatten}(\Delta{\mathbf{W}_l^k})\right)\right],
    \end{aligned}
    $%
}
\end{equation}
where $l$ represents the number of cross-attention layer of U-net, and $\mathrm{Sim}(.)$ denotes the cosine similarity calculation. To better calculate the similarity between past LoRA weight and current one, we flatten out the LoRA weight. We slow down the updating on key weights through $\mathcal{L}_{w}$, which is definitively crucial for T2I diffusion model to maintain the knowledge learned from previous styles.

While $\mathcal{L}_{w}$ regularizes the diffusion model from the perspective of weight, it overlooks the forgetting of representation from past styles, which is an indispensable aspect to overcome catastrophic forgetting. Moreover, although shared-LoRA explores the task-shared global representation of styles, the task-specific representation for each customized style is neglected. We thus consider the unique knowledge of representation from each style, and introduce a feature representation regularization in shared-LoRA. Specifically, to transfer the style knowledge from past denoising model $\epsilon^{k-1}_\theta$ to the current one $\epsilon^{k}_\theta$ and well align feature representation, we input a set of past style prompts $\mathcal{P}^{k-1} = \{p_i^j\}_{j=1}^{k-1}$ into both past and current model. Subsequently, we can generate two noise graphs $\hat{z}_i^{j}$ and $\hat{z}_i^{k}$ from past and current denoising models, respectively. We consider noise graph $\hat{z}_i^{j}$ generated from $\epsilon^{k-1}_\theta$ as a pseudo noise graph for past encountered styles, which implicitly provides strong semantic guidance for current model to maintain prior knowledge. Therefore, we devise a loss function to transfer the knowledge of pseudo noise graph $\hat{z}_i^{k-1}$ to current denoising model $\epsilon^{k}_\theta$. Formally, the loss on the $k$-th style generation task can be computed as the follows:
\begin{align}
        \label{eq: lf}
        \mathcal{L}_{f}=\frac{1}{K}\frac{1}{\mathcal{B}}\sum_{j=1}^K\sum_{i=1}^{\mathcal{B}}\rho(\hat{z}_i^{j}/\tau)\mathrm{log}(\frac{\rho(\hat{z}_i^{j}/\tau)}{\rho(\hat{z}_i^k/\tau)}).
\end{align} 

The knowledge from old styles can be transferred by $\mathcal{L}_f$, which regularizes the T2I diffusion model from the aspect of features. 

Overall, the dual regularization loss for shared-LoRA is:
\begin{align}
        \label{eq: dr}
        \mathcal{L}_{\mathrm{DR}}=\lambda_1\mathcal{L}_{w} + \lambda_2\mathcal{L}_{f},
\end{align} 
where $\lambda_1$ and $\lambda_2$ are the hyperparameters to trade-off the regularization between LoRA weights and features.

\renewcommand{\algorithmicrequire}{\textbf{Input:}}
\renewcommand{\algorithmicensure}{\textbf{Output:}}

\begin{algorithm}[t]			
    	\caption{Optimization pipeline of Our MuseumMaker.} 
	\label{alg: optimization}
	\begin{algorithmic}[1]
		\REQUIRE VAE image encoder $\mathcal{F}$, text encoder $\psi$, diffusion model $ \epsilon_{\theta} $, dataset $\mathbf{D}^K=\{\mathcal{D}^{1}_{n_k},\mathcal{D}^{2}_{n_k},\cdots,\mathcal{D}^{K}_{n_k}\}$, past style prompts set $\mathcal{P}^{K-1} = \{p_i^j\}_{j=1}^{k-1}$, number epoch  $E$, batch size $\mathcal{B}$, number of tasks ${K}$;
  \FOR{$k=1,2,\cdots, K$}
    \IF{$k=1$ }
    \STATE Initialize LoRA weight $\Delta{\mathbf{W}^k}$;
    \ELSE
    \STATE Load old LoRA weight $\Delta{\mathbf{W}}^{k-1}$;
    \ENDIF
  \STATE Initialize token embedding $\mathcal{V}^k_{*}$
  \STATE Calculate average image latent features $\bar{z}^k_{t}=\frac1{n_k}\sum_{j=1}^{n_k}\mathcal{F}(x_j^k)$ , $x_j^k \in \mathcal{D}^{k}_{n_k} ;$
  \FOR{$e=1,2,\cdots, E$}
    \STATE Random select $\mathcal{B}$ samples from $\mathcal{D}^{k}_{n_k}$
    \FOR {$i=1,2,\cdots, \mathcal{B}$}

        \STATE  Calculate single image latent features $z_{i,t}^k = \mathcal{F}(x_i^k)$;
        \STATE Calculate text embedding $c_i^k = \psi(p_i^k)$;

        \STATE  Generate noise graphs $\hat{x}_{i,t}^{k} = \epsilon_{\theta,t}(z_{i,t}^k|c_i^{k},t)$, $\bar{x}_t^{k} = \epsilon_{\theta,t}(\bar{z_t}^{k}|c_i^{k},t)$ corresponding to $z_{i,t}^K$ and $\bar{z}^k_{t}$;
        \STATE Calculate \emph{style distillation loss} $\mathcal{L}_{\mathrm{SDL}}$ by Eq.~(\ref{eq:eq9});
        \IF{$k>1$ }
            \FOR{$j = 1,2,....,k-1$ }
                \STATE Compute pseudo noise graph  $\hat{z}_i^{j} = \epsilon^{k-1}_{\theta}(p_i^j)$;
                \STATE Compute the dual regularization loss $\mathcal{L}_{\mathrm{DR}}$ by Eq.~(\ref{eq: dr});
            \ENDFOR
        \ENDIF
        \STATE Optimize diffusion model $ \epsilon^k_{\theta,t} $ and token embedding $\mathcal{V}^k_{*}$  by Eq.~(\ref{eq: eq11});

    \ENDFOR
  \ENDFOR 
  \ENDFOR

 \RETURN Current token embedding $\mathcal{V}_*^K$, and current LoRA weight $\Delta{\mathbf{W}}^K$    
\end{algorithmic} 

\end{algorithm}
 	\vspace{-10pt}


\subsection{Task-wise Token Learning}\label{chap:4.2} 
Deploying the above DR-LoRA module for continual style customization presents a promising method for sharing knowledge across diverse stylistic domains. However, when the T2I diffusion model encounters a massive continuous data stream with various styles, the DR-LoRA is limited to capture the distinct feature of each style. Additionally, as the diffusion model acquires different stylistic features, it may generate images intertwined with multiple styles, resulting in unsatisfactory generative performance. To address these challenges, we expend the model parameters of text embeddings minimally to enable the continual style learning. Furthermore, we devise unique token embeddings for each style to distinguish stylistic features from different style customization tasks. To be specific, prior works \cite{b4}\cite{b5} have extensively explored token training, where a trainable token embedding can efficiently inject a special style to diffusion model for further images generation. Therefore, we develop a \underline{T}ask-wise \underline{T}oken \underline{L}earning (TTL) module based on the textual inversion method. TTL module learns an independent token for each style customization task, and only stores the fine-tuned token parameters to maintain the prior knowledge, which take up extremely little memory. With a unique token corresponding to a distinctive style, the T2I diffusion model is able to capture unique features in a variety of styles. Formally, this module for the $k$-th task can be represented as:
\begin{align}
        \label{eq: eq3}
       v^k_*=\frac1{\mathcal{B}}\sum_{i=1}^\mathcal{B}\arg\min_{v^*}\mathbb{E}_{z_i^k,p^k_i,c_i^k,\varepsilon,t}\Big[\|\varepsilon-\epsilon_\theta(z^k_t,t,c_\theta(p_i^k))\|_2^2\Big],
\end{align}
where $v_*^k$ is the learnable token embedding, and $p_i^k \in \mathcal{P}^k$ denotes the user-provided conditioned style images of the $k$-th task. We combine the text description $v^k_*$ with the conditioned stylistic images by Eq.~\eqref{eq: eq3}. The T2I diffusion model fuses the features of conditioned model, which can be triggered to generate specific style images when the corresponding $v^k_*$ is input.

However, Textual Inversion~\cite{b4} overlooks the independence of cross-attention layers at different resolutions during denoising process. Solely training the input token may not sufficiently unfold the embedding space. Inspired by this observation, we extend the input token embedding by introducing multiple independent tokens, which are learned across different cross-attention layers of the U-Net. We denote the extended $v_*^k$ as $\mathcal{V}_*^k = \{v^k_1, v^k_2, ..., v^k_L\}$, where $k$ represents the corresponding task index, $L$ represents the number of layers of cross-attention. The trainable tokens set of $K$ tasks can be written as $\boldsymbol{V}_*^K = \{\mathcal{V}_*^1, \mathcal{V}_*^2, ..., \mathcal{V}_*^K \}$. Consequently, the Eq.~\eqref{eq: eq3} for $K$ tasks can be rewritten as:
\begin{equation}
\resizebox{0.47\textwidth}{!}{%
    $\displaystyle
    \begin{aligned}
        \label{eq: eq4}
        \boldsymbol{V}_*^K=\frac1{\mathcal{B}}\sum_{k=1}^K\sum_{i=1}^\mathcal{B}\arg\min_v\mathbb{E}_{z_i^k,p^k_i,c_i^k,\varepsilon,t}\Big[\|\varepsilon-\epsilon_\theta(z^k_{i,t},t,c_\theta(p_i^k))\|_2^2\Big],
    \end{aligned}
    $%
}
\end{equation}
where each task obtains a extended token. We effectively extend the dimension of the token embedding space by training an independent token embedding for each cross-attention layer, which enhances the final generation quality. After training, we store a special token embedding for each task to further mitigate catastrophic forgetting during continual style customization.

In summary, the overall pipeline optimization for our proposed MuseumMaker method is illustrated in Algorithm 1. The loss function constrained by prior knowledge to enable continual learning can be formulated as:
\begin{align}
        \label{eq: eq10}
       \mathcal{L}_{c}=\alpha\mathcal{L}_{\mathrm{SDL}} + \beta\mathcal{L}_{\mathrm{DR}},
\end{align}
where $\alpha$ and $\beta$ are manually set hyperparameters. The total optimization objective can be formulated as follows:
\begin{align}
        \label{eq: eq11}
       \mathcal{L}_{\mathrm{Overall}}=\sum_{k=1}^K(\mathcal{L}^k_{\mathrm{SD}}+ \alpha\mathcal{L}_{\mathrm{SDL}} + \beta\mathcal{L}_{\mathrm{DR}}).
\end{align}

Regarding $\mathcal{L}_{\mathrm{Overall}}$, our MuseumMaker systematically aggregates a continual collection of styles, empowering users to draw artworks of diverse styles and build their own customized museum.
\begin{figure*}[p]
	\centering
        \includegraphics[width=515pt,height=\textheight,keepaspectratio]{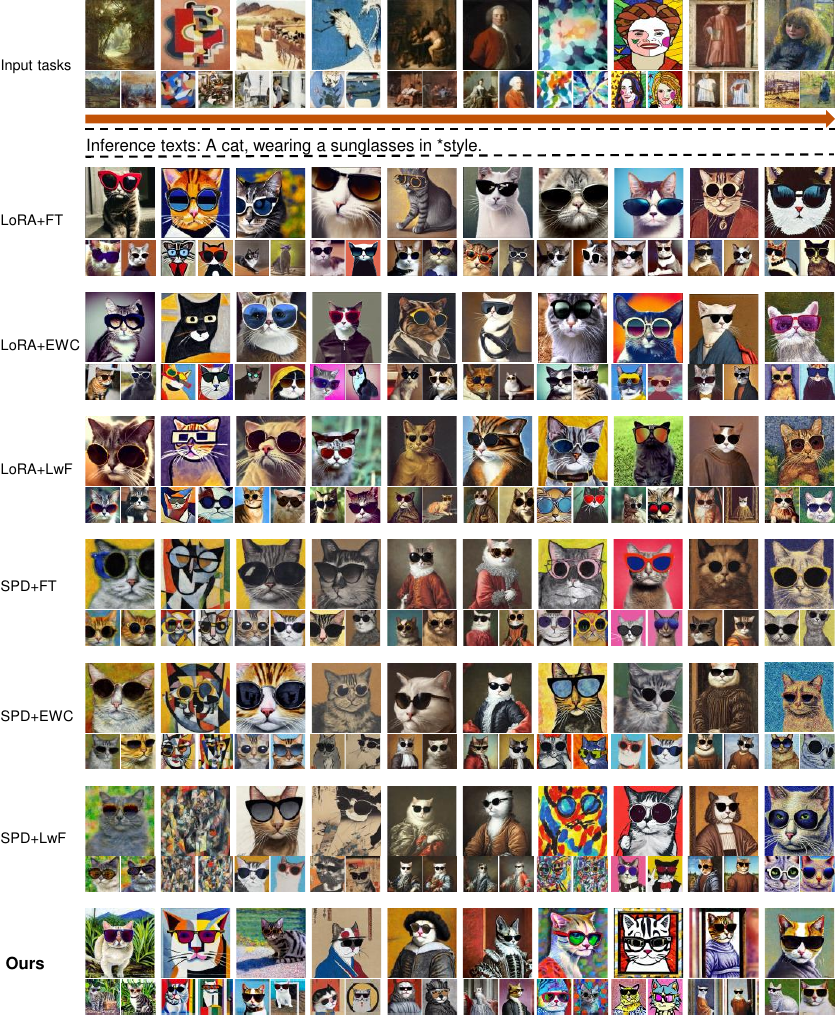} 
 	\vspace{-5pt}
	\caption{Qualitative comparison between our method with competing methods in 10 tasks continual style adaption setting, where the first two rows represent the stylistic dataset and prompts provides users, and the rest results denote the image generated by each method with the same text prompt, \emph{i.e., a cat, wearing a sunglasses in *style}.} 
	\label{fig: experiment1}
\end{figure*}

\begin{figure*}[t]
	\centering
     \hspace{-0cm} 
	\includegraphics[width=517pt,height=430pt]
	{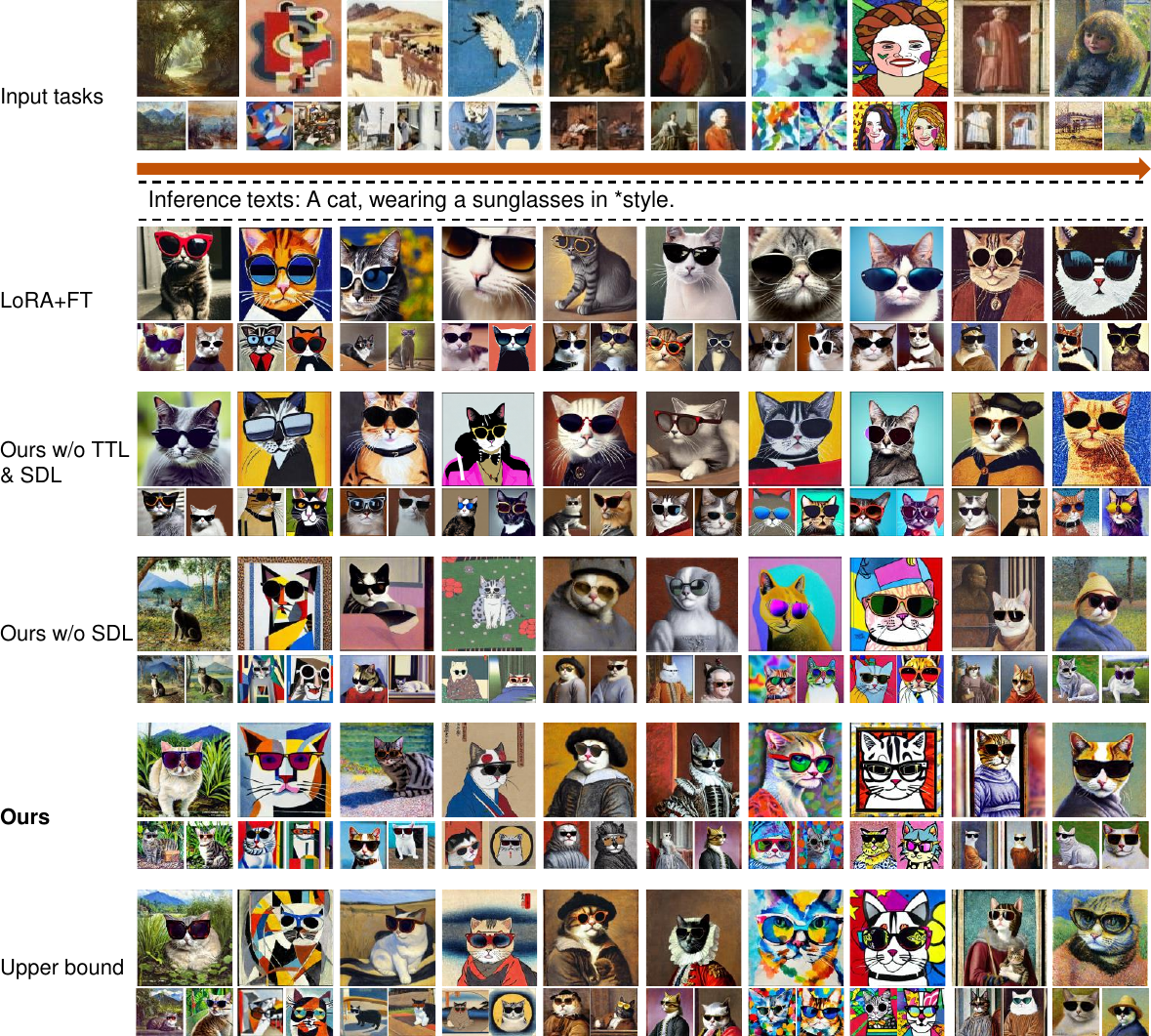}
 	\vspace{-5pt}
	\caption{Qualitative comparison of our ablation studies, where we evaluate the contribution of each module we proposed. We denotes the task-wise token learning module as TTL and style distillation loss as SDL in the fig, respectively. We input the same text prompt as we do in comparison experiments. The upper bound setting stores a learnable token embedding and LoRA weight for each style } 
	\label{fig: experiment2}
\end{figure*}

\section{Experiment}
\subsection{Datasets and Evaluation}
We present comparison experiments and ablation studies on Wikiart\cite{b69} to illustrate the efficacy of our MuseumMaker. Due to the particularity of imges generation, we utilize three kinds of metrics (\emph{i.e.}, style loss\cite{b10}, FID\cite{fid} and CLIP score\cite{clip_score}) to measure the superior performance of our proposed MuseumMaker.

\textbf{Datasets}: WikiArt consists of a diverse and rich artwork dataset of diferent styles, which encompasses artworks from 61 genres and localizes in 8 languages. The artworks from WikiArt is sourced from museums, universities, city halls, and other municipal buildings spanning over 100 countries.
For evaluation, we select 10 datasets representing different artistic styles in continual learning manner, as follow: (1) impressionism, (2) cubism, (3) realism, (4) ukiyo, (5) baroque, (6) rococo, (7) expressionism, (8) pop art, (9) renaissance, (10) pointillism. Each dataset comprises 8-10 images, showcasing distinct distributions of artistic styles. For ease of reference, we abbreviate each style to the first three letters.

\begin{table*}[htbp]
\setlength{\tabcolsep}{1.0mm}
\renewcommand\arraystretch{1.5}
\normalsize
\centering
  \caption{Continual style adaptation generation comparison between ours with state-of-the-arts method (\emph{e.g.,}LoRA\cite{b8}+LWF\cite{lwf}, LoRA\cite{b8}+EWC\cite{b11}, SPD\cite{b10}+LWF\cite{lwf}, SPD\cite{b10}+EWC\cite{b11}) in terms of FID and CLIP score. Methods with the best and runner-up performance are marked as bolded {\color[HTML]{FE0000} \textbf{red}} and {\color[HTML]{3531FF} \textbf{blue}}, respectively.}
  \scalebox{0.66}{
           \begin{tabular}{l|cccccccccccc|cccccccccccc}
\hline
\multicolumn{1}{c|}{}                                     & \multicolumn{12}{c|}{FID$\Downarrow$}                                                                                                                                                                                                                                                                                                                                                                                                                                                                                                                  & \multicolumn{12}{c}{CLIP Score$\Uparrow$}                                                                                                                                                                                                                                                                                                                                                                                                                                                                                                             \\
\multicolumn{1}{c|}{\multirow{-2}{*}{Comparison Methods}} & imp                                   & cub                                   & rea                                   & uki                                   & bar                                   & roc                                   & exp                                   & pop                                   & ren                                   & \multicolumn{1}{c|}{poi}                                   & \textbf{Ave.}                                                 & \textbf{Imp.}                         & imp                                   & cub                                   & rea                                   & uki                                   & bar                                   & roc                                   & exp                                   & pop                                   & ren                                   & \multicolumn{1}{c|}{poi}                                   & \textbf{Ave.}                                                 & \textbf{Imp.}                          \\ \hline
LoRA+FT                                                   & 334.7                                 & 371.8                                 & 399.2                                 & 381.2                                 & 406.8                                 & 325.0                                 & 449.4                                 & 401.6                                 & 351.2                                 & \multicolumn{1}{c|}{382.7}                                 & \cellcolor[HTML]{E7E6E6}\textbf{380.4}                        & \cellcolor[HTML]{E7E6E6}\textbf{$\Uparrow$66.9} & 51.78                                 & 68.66                                 & 53.83                                 & 54.05                                 & 59.38                                 & 53.43                                 & 58.31                                 & 50.97                                 & 59.17                                 & \multicolumn{1}{c|}{55.30}                                 & \cellcolor[HTML]{E7E6E6}\textbf{56.49}                        & \cellcolor[HTML]{E7E6E6}\textbf{$\Uparrow$11.77} \\
LoRA+LWF                                                  & {\color[HTML]{0000FF} \textbf{328.1}} & 365.9                                 & 398.4                                 & 386.7                                 & 411.9                                 & 331.7                                 & 428.4                                 & 408.7                                 & 360.1                                 & \multicolumn{1}{c|}{381.4}                                 & \cellcolor[HTML]{E7E6E6}\textbf{380.1}                        & \cellcolor[HTML]{E7E6E6}\textbf{$\Uparrow$66.7} & 52.09                                 & 69.72                                 & 54.05                                 & 53.70                                 & 58.17                                 & 52.03                                 & 65.77                                 & 49.36                                 & 56.69                                 & \multicolumn{1}{c|}{55.10}                                 & \cellcolor[HTML]{E7E6E6}\textbf{56.67}                        & \cellcolor[HTML]{E7E6E6}\textbf{$\Uparrow$11.59} \\
LoRA+EWC                                                  & 335.7                                 & 377.1                                 & 398.3                                 & 380.6                                 & 405.1                                 & 321.9                                 & 448.4                                 & 404.9                                 & 351.1                                 & \multicolumn{1}{c|}{383.9}                                 & \cellcolor[HTML]{E7E6E6}\textbf{380.7}                        & \cellcolor[HTML]{E7E6E6}\textbf{$\Uparrow$67.2} & 51.99                                 & 69.01                                 & 53.88                                 & 53.81                                 & 59.47                                 & 53.87                                 & 58.27                                 & 50.92                                 & 59.02                                 & \multicolumn{1}{c|}{55.48}                                 & \cellcolor[HTML]{E7E6E6}\textbf{56.57}                        & \cellcolor[HTML]{E7E6E6}\textbf{$\Uparrow$11.69} \\
SPD+FT                                                    & 345.4                                 & {\color[HTML]{FF0000} \textbf{262.2}} & 396.3                                 & 315.9                                 & {\color[HTML]{0000FF} \textbf{394.6}} & 287.3                                 & 386.0                                 & 358.2                                 & {\color[HTML]{0000FF} \textbf{314.7}} & \multicolumn{1}{c|}{361.8}                                 & \cellcolor[HTML]{E7E6E6}\textbf{342.2}                        & \cellcolor[HTML]{E7E6E6}\textbf{$\Uparrow$28.8} & 58.77                                 & {\color[HTML]{0000FF} \textbf{84.07}} & 53.71                                 & 70.15                                 & 62.23                                 & {\color[HTML]{0000FF} \textbf{64.21}} & 70.93                                 & 56.05                                 & 68.35                                 & \multicolumn{1}{c|}{56.52}                                 & \cellcolor[HTML]{E7E6E6}\textbf{64.50}                        & \cellcolor[HTML]{E7E6E6}\textbf{$\Uparrow$3.76}  \\
SPD+LWF                                                   & 331.1                                 & 378.3                                 & {\color[HTML]{0000FF} \textbf{395.3}} & 329.0                                 & 394.8                                 & {\color[HTML]{FF0000} \textbf{271.6}} & {\color[HTML]{FF0000} \textbf{264.2}} & {\color[HTML]{0000FF} \textbf{328.4}} & 327.0                                 & \multicolumn{1}{c|}{365.7}                                 & \cellcolor[HTML]{E7E6E6}\textbf{338.5}                        & \cellcolor[HTML]{E7E6E6}\textbf{$\Uparrow$25.1} & {\color[HTML]{0000FF} \textbf{59.59}} & 82.65                                 & 53.94                                 & {\color[HTML]{0000FF} \textbf{75.86}} & {\color[HTML]{0000FF} \textbf{62.26}} & {\color[HTML]{FF0000} \textbf{69.72}} & {\color[HTML]{FF0000} \textbf{85.92}} & {\color[HTML]{0000FF} \textbf{59.10}} & {\color[HTML]{FF0000} \textbf{68.59}} & \multicolumn{1}{c|}{55.34}                                 & \cellcolor[HTML]{E7E6E6}{\color[HTML]{0000FF} \textbf{67.30}} & \cellcolor[HTML]{E7E6E6}\textbf{$\Uparrow$0.96}  \\
SPD+EWC                                                   & 345.9                                 & 273.7                                 & 396.8                                 & {\color[HTML]{0000FF} \textbf{298.2}} & 394.9                                 & 284.6                                 & {\color[HTML]{0000FF} \textbf{356.3}} & 347.4                                 & {\color[HTML]{FF0000} \textbf{313.4}} & \multicolumn{1}{c|}{{\color[HTML]{0000FF} \textbf{363.5}}} & \cellcolor[HTML]{E7E6E6}{\color[HTML]{0000FF} \textbf{337.5}} & \cellcolor[HTML]{E7E6E6}\textbf{$\Uparrow$24.0} & 58.72                                 & {\color[HTML]{FF0000} \textbf{84.73}} & {\color[HTML]{0000FF} \textbf{54.70}} & 71.22                                 & 62.10                                 & 63.92                                 & {\color[HTML]{0000FF} \textbf{73.75}} & 57.72                                 & {\color[HTML]{0000FF} \textbf{68.50}} & \multicolumn{1}{c|}{{\color[HTML]{0000FF} \textbf{57.69}}} & \cellcolor[HTML]{E7E6E6}\textbf{65.31}                        & \cellcolor[HTML]{E7E6E6}\textbf{$\Uparrow$2.96}  \\ \hline
\textbf{Ours}                                             & {\color[HTML]{FF0000} \textbf{250.5}} & {\color[HTML]{0000FF} \textbf{271.2}} & {\color[HTML]{FF0000} \textbf{374.9}} & {\color[HTML]{FF0000} \textbf{251.8}} & {\color[HTML]{FF0000} \textbf{348.2}} & {\color[HTML]{0000FF} \textbf{286.3}} & 373.1                                 & {\color[HTML]{FF0000} \textbf{312.1}} & 319.5                                 & \multicolumn{1}{c|}{{\color[HTML]{FF0000} \textbf{347.2}}} & \cellcolor[HTML]{E7E6E6}{\color[HTML]{FF0000} \textbf{313.5}} & \cellcolor[HTML]{E7E6E6}\textbf{-}    & {\color[HTML]{FF0000} \textbf{68.86}} & 82.99                                 & {\color[HTML]{FF0000} \textbf{60.18}} & {\color[HTML]{FF0000} \textbf{79.21}} & {\color[HTML]{FF0000} \textbf{67.15}} & 62.98                                 & 70.56                                 & {\color[HTML]{FF0000} \textbf{59.69}} & 63.46                                 & \multicolumn{1}{c|}{{\color[HTML]{FF0000} \textbf{67.53}}} & \cellcolor[HTML]{E7E6E6}{\color[HTML]{FF0000} \textbf{68.26}} & \cellcolor[HTML]{E7E6E6}\textbf{-}     \\ \hline
Upper Bound                                               & 249.5                                 & 254.6                                 & 350.0                                 & 236.5                                 & 317.3                                 & 279.8                                 & 245.0                                 & 335.6                                 & 260.4                                 & \multicolumn{1}{c|}{327.5}                                 & \cellcolor[HTML]{E7E6E6}\textbf{285.6}                        & \cellcolor[HTML]{E7E6E6}\textbf{-}    & 74.35                                 & 84.20                                 & 67.35                                 & 81.84                                 & 71.94                                 & 68.40                                 & 82.32                                 & 60.37                                 & 75.46                                 & \multicolumn{1}{c|}{70.71}                                 & \cellcolor[HTML]{E7E6E6}\textbf{73.69}                        & \cellcolor[HTML]{E7E6E6}\textbf{-}     \\ \hline
\end{tabular}

\label{tab:tab1}}
\vspace{-5pt}
\end{table*}

\begin{table}[t]
\centering
\setlength{\tabcolsep}{1.0mm}
\renewcommand{\arraystretch}{1.5}
\caption{ Continual style adaptation generation comparison between ours with state-of-the-arts method (\emph{e.g.,}LoRA\cite{b8}+LWF\cite{lwf}, LoRA\cite{b8}+EWC\cite{b11}, SPD\cite{b10}+LWF\cite{lwf}, SPD\cite{b10}+EWC\cite{b11}) in terms of style loss. In order to better display the data results, all results of style loss is scaled by 100. Methods with the best and runner-up performance are marked as bolded {\color[HTML]{FE0000} \textbf{red}} and {\color[HTML]{3531FF} \textbf{blue}}, respectively.}
\scalebox{0.69}{
\begin{tabular}{l|cccccccccccc}
\hline
\multicolumn{1}{c|}{}                                     & \multicolumn{12}{c}{Style Loss$\Downarrow$}                                                                                                                                                                                                                                                                                                                                                                                                                                                                                                             \\
\multicolumn{1}{c|}{\multirow{-2}{*}{Comparison Methods}} & imp                                   & cub                                   & rea                                   & uki                                   & bar                                   & roc                                   & exp                                   & pop                                   & ren                                   & \multicolumn{1}{c|}{poi}                                   & \textbf{Ave.}                                                 & \textbf{Imp.}                          \\ \hline
LoRA+FT                                                   & 0.259                                 & 0.055                                 & 0.135                                 & 0.161                                 & 0.083                                 & 0.076                                 & 0.249                                 & 0.208                                 & 0.107                                 & \multicolumn{1}{c|}{{\color[HTML]{0000FF} \textbf{0.038}}} & \cellcolor[HTML]{E7E6E6}\textbf{0.137}                        & \cellcolor[HTML]{E7E6E6}\textbf{$\Uparrow$0.058} \\
LoRA+LWF                                                  & 0.239                                 & {\color[HTML]{FF0000} \textbf{0.045}} & 0.116                                 & 0.172                                 & 0.079                                 & 0.070                                 & 0.095                                 & 0.195                                 & 0.079                                 & \multicolumn{1}{c|}{0.045}                                 & \cellcolor[HTML]{E7E6E6}\textbf{0.114}                        & \cellcolor[HTML]{E7E6E6}\textbf{$\Uparrow$0.035} \\
LoRA+EWC                                                  & 0.233                                 & {\color[HTML]{0000FF} \textbf{0.050}} & 0.111                                 & 0.176                                 & 0.077                                 & 0.074                                 & 0.238                                 & 0.210                                 & 0.087                                 & \multicolumn{1}{c|}{{\color[HTML]{FF0000} \textbf{0.037}}} & \cellcolor[HTML]{E7E6E6}\textbf{0.129}                        & \cellcolor[HTML]{E7E6E6}\textbf{$\Uparrow$0.050} \\
SPD+FT                                                    & {\color[HTML]{FF0000} \textbf{0.025}} & 0.085                                 & 0.104                                 & 0.040                                 & {\color[HTML]{0000FF} \textbf{0.073}} & {\color[HTML]{FF0000} \textbf{0.062}} & {\color[HTML]{0000FF} \textbf{0.065}} & 0.081                                 & {\color[HTML]{FF0000} \textbf{0.073}} & \multicolumn{1}{c|}{0.084}                                 & \cellcolor[HTML]{E7E6E6}{\color[HTML]{FF0000} \textbf{0.069}} & \cellcolor[HTML]{E7E6E6}\textbf{$\Uparrow$0.010} \\
SPD+LWF                                                   & {\color[HTML]{0000FF} \textbf{0.036}} & 0.063                                 & {\color[HTML]{0000FF} \textbf{0.084}} & {\color[HTML]{FF0000} \textbf{0.035}} & {\color[HTML]{FF0000} \textbf{0.056}} & {\color[HTML]{0000FF} \textbf{0.065}} & {\color[HTML]{FF0000} \textbf{0.048}} & {\color[HTML]{FF0000} \textbf{0.072}} & 0.133                                 & \multicolumn{1}{c|}{0.223}                                 & \cellcolor[HTML]{E7E6E6}\textbf{0.081}                        & \cellcolor[HTML]{E7E6E6}\textbf{$\Uparrow$0.002} \\
SPD+EWC                                                   & 0.040                                 & 0.139                                 & 0.125                                 & 0.043                                 & 0.105                                 & 0.092                                 & 0.075                                 & {\color[HTML]{0000FF} \textbf{0.079}} & 0.093                                 & \multicolumn{1}{c|}{0.096}                                 & \cellcolor[HTML]{E7E6E6}\textbf{0.089}                        & \cellcolor[HTML]{E7E6E6}\textbf{$\Uparrow$0.010} \\ \hline
\textbf{Ours}                                             & 0.060                                 & 0.100                                 & {\color[HTML]{FF0000} \textbf{0.065}} & {\color[HTML]{0000FF} \textbf{0.039}} & 0.089                                 & 0.090                                 & 0.076                                 & 0.108                                 & {\color[HTML]{0000FF} \textbf{0.077}} & \multicolumn{1}{c|}{0.082}                                 & \cellcolor[HTML]{E7E6E6}{\color[HTML]{0000FF} \textbf{0.079}} & \cellcolor[HTML]{E7E6E6}\textbf{-}     \\ \hline
Upper Bound                                               & 0.030                                 & 0.093                                 & 0.049                                 & 0.039                                 & 0.040                                 & 0.093                                 & 0.103                                 & 0.182                                 & 0.099                                 & \multicolumn{1}{c|}{0.059}                                 & \cellcolor[HTML]{E7E6E6}\textbf{0.079}                        & \cellcolor[HTML]{E7E6E6}\textbf{-}     \\ \hline
\end{tabular}}
\label{tab: sytle_loss}
\vspace{-10pt}
\end{table}


\textbf{Evaluation}:  To fairly evaluate the generative performance of models in a continual learning scenario, we leverage hundreds of images generated from 20 prompts for quantitative evaluation, with each prompt generating 50 sets of images. In order to evaluate the similarity of style between the generated images and the original style, we introduce style loss, FID and CLIP score. The details of these three metrics are as follow:

1) \textbf{Style loss} is employed in the field of image generation to assess the stylistic similarity between two images. We take the images from the training dataset as references and calculate the style similarity between these references and all generated images, subsequently averaging the results; 2) \textbf{FID} assesses the quality of generated images by quantifying the distance between source images and generated, which considers the distribution between both source data; 3) \textbf{CLIP score} utilizes the pre-trained ViT-B/32 model from CLIP to measure the property of diffusion model. By encoding both the generated images and reference images using CLIP, their similarity in the latent space can be computed. A higher similarity score indicates a better correspondence between the generated images and reference images.

\begin{figure*}[t]
	\centering
	\includegraphics[width=7.1in]
	{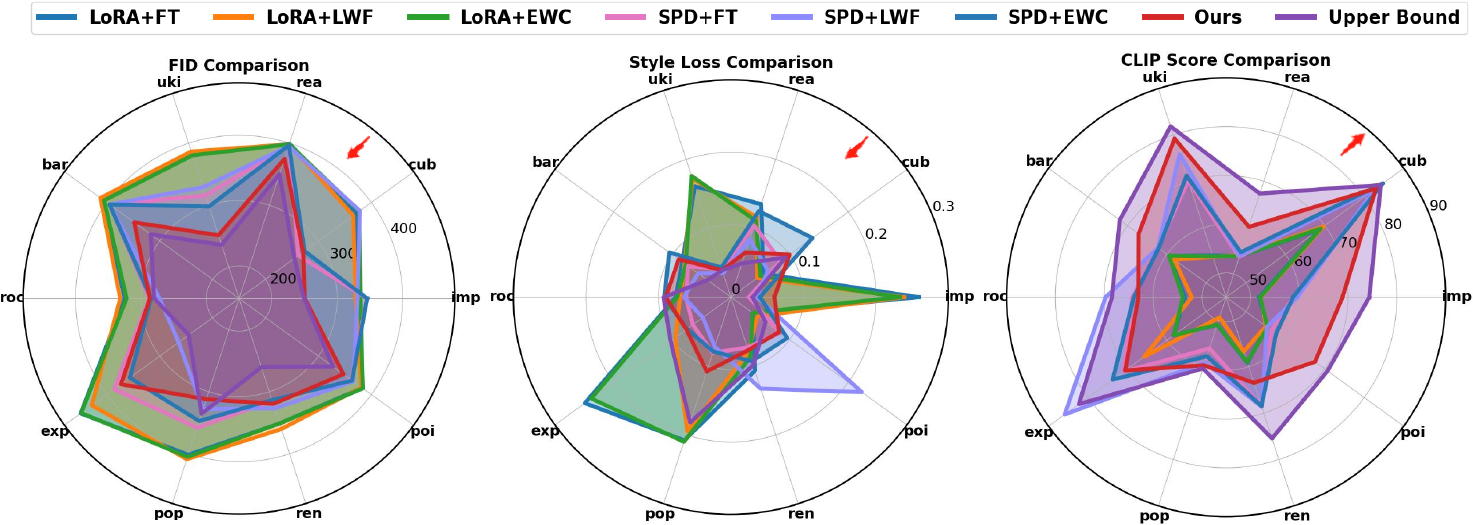}
 	\vspace{-5pt}
	\caption{Style loss, FID and CLIP score for continual style adaptation setting. The red arrow in the image indicates the direction where the score improves. Comparing with other competing method (\emph{e.g.,}  LoRA\cite{b8}+LWF, LoRA+EWC, SPD+LWF, SPD+EWC), our MuseumMaker achieves the best performance in terms of FID and CLIP score.} 
	\label{fig: table1}
\end{figure*}

\subsection{ Implementation Details and Baselines}
In this work, we conduct continual style customization experiments on our proposed MuseumMaker, LoRA\cite{b8} and Specialist Diffusion (SPD)\cite{b10}. For the continual style customization experiment, we conduct the experiment on the 10 style datasets, ensuring that the previous style is not available during the training process. 
To assess the effectiveness of our continual style customization method in contrast to existing approaches, we deploy Fine-tuning (FT), as well as two classic continual learning methods, namely EWC \cite{b11} and LWF \cite{lwf}, for LoRA and SPD. These methods serve as benchmarks for comparison with our approach.
Furthermore, we store a learnable token embedding and LoRA weight for each style as our upper bound. During the learning phase for each style, we employ 1000 training steps for LoRA baselines method, 100 training epoch for SPD baselines methods, 1000 training steps for ours. All the methods are implied with a batch size of 1. To achieve desirable generation performance, we set the learning rate to $1 \times 10^{-5}$ for LoRA baselines method, $1 \times 10^{-6}$ for SPD baselines methods and $1 \times 10^{-5}$ for our method. Additionally, the hyperparameters $\lambda_1$ and $\lambda_2$ are set 0.8 and 1.0 in Eq.~(\ref{eq: dr}),  $\alpha$ and $\beta$ in Eq.~(\ref{eq: eq11}) are set to 0.8 and 1.5. During the inference stage, we apply a DDIM sampler with 50 steps for all methods. All experiments are conducted on a single NVIDIA RTX A6000 GPU, with all generated images having a resolution of $512 \times 512$ pixels and based on Stable Diffusion v1.5.

\begin{table*}[htbp]
\setlength{\tabcolsep}{1.0mm}
\renewcommand\arraystretch{1.5}
\normalsize
\centering
  \caption{Comparison experiments of our ablation studies, where TTL and SDL represent task-wise token learning module and style distillation loss module, respectively. Methods except for the upper bound with the best and runner-up performance are marked as bolded {\color[HTML]{FE0000} \textbf{red}} and {\color[HTML]{3531FF} \textbf{blue}}, respectively. }
  \scalebox{0.655}{
           \begin{tabular}{l|cccccccccccc|cccccccccccc}
\hline
\multicolumn{1}{c|}{}                                     & \multicolumn{12}{c|}{FID$\Downarrow$}                                                                                                                                                                                                                                                                                                                                                                                                                                                                                                                  & \multicolumn{12}{c}{CLIP Score$\Uparrow$}                                                                                                                                                                                                                                                                                                                                                                                                                                                                                                             \\
\multicolumn{1}{c|}{\multirow{-2}{*}{Comparison Methods}} & imp                                   & cub                                   & rea                                   & uki                                   & bar                                   & roc                                   & exp                                   & pop                                   & ren                                   & \multicolumn{1}{c|}{poi}                                   & \textbf{Ave.}                                                 & \textbf{Imp.}                         & imp                                   & cub                                   & rea                                   & uki                                   & bar                                   & roc                                   & exp                                   & pop                                   & ren                                   & \multicolumn{1}{c|}{poi}                                   & \textbf{Ave.}                                                 & \textbf{Imp.}                          \\ \hline
LoRA+FT                                                   & 334.7                                 & 371.8                                 & 399.2                                 & 381.2                                 & 406.8                                 & 325.0                                 & 449.4                                 & 401.6                                 & 351.2                                 & \multicolumn{1}{c|}{382.7}                                 & \cellcolor[HTML]{E7E6E6}\textbf{380.4}                        & \cellcolor[HTML]{E7E6E6}\textbf{$\Uparrow$66.9} & 51.78                                 & 68.66                                 & 53.83                                 & 54.05                                 & 59.38                                 & 53.43                                 & 58.31                                 & 50.97                                 & 59.17                                 & \multicolumn{1}{c|}{55.30}                                 & \cellcolor[HTML]{E7E6E6}\textbf{56.49}                        & \cellcolor[HTML]{E7E6E6}\textbf{$\Uparrow$11.77} \\
Ours w/o TTL \& SDL                                      & 332.6                                 & 366.0                                 & 401.7                                 & 382.4                                 & 406.4                                 & 318.8                                 & 435.9                                 & 406.5                                 & 348.2                                 & \multicolumn{1}{c|}{387.9}                                 & \cellcolor[HTML]{E7E6E6}\textbf{378.6}                        & \cellcolor[HTML]{E7E6E6}\textbf{$\Uparrow$65.2} & 51.68                                 & 68.93                                 & 53.87                                 & 53.35                                 & 58.43                                 & 52.10                                 & 65.87                                 & 49.45                                 & 56.23                                 & \multicolumn{1}{c|}{55.11}                                 & \cellcolor[HTML]{E7E6E6}\textbf{56.50}                        & \cellcolor[HTML]{E7E6E6}\textbf{$\Uparrow$11.76} \\
Ours w/o SDL                                              & {\color[HTML]{FF0000} \textbf{233.8}} & {\color[HTML]{0000FF} \textbf{321.1}} & {\color[HTML]{0000FF} \textbf{382.0}} & {\color[HTML]{0000FF} \textbf{284.1}} & {\color[HTML]{0000FF} \textbf{377.8}} & {\color[HTML]{FF0000} \textbf{285.8}} & {\color[HTML]{FF0000} \textbf{357.0}} & {\color[HTML]{0000FF} \textbf{314.7}} & {\color[HTML]{0000FF} \textbf{328.8}} & \multicolumn{1}{c|}{{\color[HTML]{FF0000} \textbf{340.4}}} & \cellcolor[HTML]{E7E6E6}{\color[HTML]{0000FF} \textbf{322.5}} & \cellcolor[HTML]{E7E6E6}\textbf{$\Uparrow$9.1}  & {\color[HTML]{FF0000} \textbf{74.23}} & {\color[HTML]{0000FF} \textbf{82.26}} & {\color[HTML]{FF0000} \textbf{65.16}} & {\color[HTML]{0000FF} \textbf{77.13}} & {\color[HTML]{0000FF} \textbf{66.96}} & {\color[HTML]{FF0000} \textbf{65.00}} & {\color[HTML]{FF0000} \textbf{73.02}} & {\color[HTML]{0000FF} \textbf{58.92}} & {\color[HTML]{FF0000} \textbf{68.08}} & \multicolumn{1}{c|}{{\color[HTML]{FF0000} \textbf{70.25}}} & \cellcolor[HTML]{E7E6E6}{\color[HTML]{FF0000} \textbf{70.10}} & \cellcolor[HTML]{E7E6E6}\textbf{$\Downarrow$-1.84} \\ \hline
\textbf{Ours}                                             & {\color[HTML]{0000FF} \textbf{250.5}} & {\color[HTML]{FF0000} \textbf{271.2}} & {\color[HTML]{FF0000} \textbf{374.9}} & {\color[HTML]{FF0000} \textbf{251.8}} & {\color[HTML]{FF0000} \textbf{348.2}} & {\color[HTML]{0000FF} \textbf{286.3}} & {\color[HTML]{0000FF} \textbf{373.1}} & {\color[HTML]{FF0000} \textbf{312.1}} & {\color[HTML]{FF0000} \textbf{319.5}} & \multicolumn{1}{c|}{{\color[HTML]{0000FF} \textbf{347.2}}} & \cellcolor[HTML]{E7E6E6}{\color[HTML]{FF0000} \textbf{313.5}} & \cellcolor[HTML]{E7E6E6}\textbf{-}    & {\color[HTML]{0000FF} \textbf{68.86}} & {\color[HTML]{FF0000} \textbf{82.99}} & {\color[HTML]{0000FF} \textbf{60.18}} & {\color[HTML]{FF0000} \textbf{79.21}} & {\color[HTML]{FF0000} \textbf{67.15}} & {\color[HTML]{0000FF} \textbf{62.98}} & {\color[HTML]{0000FF} \textbf{70.56}} & {\color[HTML]{FF0000} \textbf{59.69}} & {\color[HTML]{0000FF} \textbf{63.46}} & \multicolumn{1}{c|}{{\color[HTML]{0000FF} \textbf{67.53}}} & \cellcolor[HTML]{E7E6E6}{\color[HTML]{0000FF} \textbf{68.26}} & \cellcolor[HTML]{E7E6E6}\textbf{-}     \\ \hline
Upper Bound                                               & 249.5                                 & 254.6                                 & 350.0                                 & 236.5                                 & 317.3                                 & 279.8                                 & 245.0                                 & 335.6                                 & 260.4                                 & \multicolumn{1}{c|}{327.5}                                 & \cellcolor[HTML]{E7E6E6}\textbf{285.6}                        & \cellcolor[HTML]{E7E6E6}\textbf{-}    & 74.35                                 & 84.20                                 & 67.35                                 & 81.84                                 & 71.94                                 & 68.40                                 & 82.32                                 & 60.37                                 & 75.46                                 & \multicolumn{1}{c|}{70.71}                                 & \cellcolor[HTML]{E7E6E6}\textbf{73.69}                        & \cellcolor[HTML]{E7E6E6}\textbf{-}     \\ \hline
\end{tabular}
\label{tab:tab2}}
\vspace{-5pt}
\end{table*}

\begin{table}[t]
\centering
\setlength{\tabcolsep}{1.0mm}
\renewcommand{\arraystretch}{1.5}
\caption{ Comparison experiments of our ablation studies, where TTL and SDL represent task-wise token learning module and style distillation loss module, respectively. In order to better display the data results, all results of style loss is scaled by 100. Methods with the best and runner-up performance are marked as bolded {\color[HTML]{FE0000} \textbf{red}} and {\color[HTML]{3531FF} \textbf{blue}}, respectively.}
\scalebox{0.67}{
\begin{tabular}{l|cccccccccccc}
\hline
\multicolumn{1}{c|}{}                                     & \multicolumn{12}{c}{Style Loss$\Downarrow$}                                                                                                                                                                                                                                                                                                                                                                                                                                                                                                             \\
\multicolumn{1}{c|}{\multirow{-2}{*}{Comparison Methods}} & imp                                   & cub                                   & rea                                   & uki                                   & bar                                   & roc                                   & exp                                   & pop                                   & ren                                   & \multicolumn{1}{c|}{poi}                                   & \textbf{Ave.}                                                 & \textbf{Imp.}                          \\ \hline
LoRA+FT                                                   & 0.259                                 & {\color[HTML]{0000FF} \textbf{0.055}} & 0.135                                 & 0.161                                 & {\color[HTML]{0000FF} \textbf{0.083}} & {\color[HTML]{0000FF} \textbf{0.076}} & 0.249                                 & 0.208                                 & 0.107                                 & \multicolumn{1}{c|}{{\color[HTML]{FF0000} \textbf{0.038}}} & \cellcolor[HTML]{E7E6E6}\textbf{0.137}                        & \cellcolor[HTML]{E7E6E6}\textbf{$\Uparrow$0.059} \\
Ours w/o TTL \& SDL                                      & {\color[HTML]{FF0000} \textbf{0.030}} & {\color[HTML]{FF0000} \textbf{0.047}} & 0.116                                 & 0.168                                 & {\color[HTML]{FF0000} \textbf{0.073}} & {\color[HTML]{FF0000} \textbf{0.062}} & {\color[HTML]{0000FF} \textbf{0.091}} & 0.186                                 & {\color[HTML]{0000FF} \textbf{0.083}} & \multicolumn{1}{c|}{{\color[HTML]{0000FF} \textbf{0.044}}} & \cellcolor[HTML]{E7E6E6}\textbf{0.090}                        & \cellcolor[HTML]{E7E6E6}\textbf{$\Uparrow$0.044} \\
Ours w/o SDL                                              & 0.062                                 & 0.083                                 & {\color[HTML]{0000FF} \textbf{0.068}} & {\color[HTML]{0000FF} \textbf{0.042}} & 0.083                                 & 0.105                                 & 0.094                                 & {\color[HTML]{0000FF} \textbf{0.161}} & 0.111                                 & \multicolumn{1}{c|}{0.053}                                 & \cellcolor[HTML]{E7E6E6}{\color[HTML]{0000FF} \textbf{0.086}} & \cellcolor[HTML]{E7E6E6}\textbf{$\Uparrow$0.007} \\ \hline
\textbf{Ours}                                             & {\color[HTML]{0000FF} \textbf{0.060}} & 0.100                                 & {\color[HTML]{FF0000} \textbf{0.065}} & {\color[HTML]{FF0000} \textbf{0.039}} & 0.089                                 & 0.090                                 & {\color[HTML]{FF0000} \textbf{0.076}} & {\color[HTML]{FF0000} \textbf{0.108}} & {\color[HTML]{FF0000} \textbf{0.077}} & \multicolumn{1}{c|}{0.082}                                 & \cellcolor[HTML]{E7E6E6}{\color[HTML]{FF0000} \textbf{0.079}} & \cellcolor[HTML]{E7E6E6}\textbf{-}     \\ \hline
Upper Bound                                               & 0.030                                 & 0.093                                 & 0.049                                 & 0.039                                 & 0.040                                 & 0.093                                 & 0.103                                 & 0.182                                 & 0.099                                 & \multicolumn{1}{c|}{0.059}                                 & \cellcolor[HTML]{E7E6E6}\textbf{0.079}                        & \cellcolor[HTML]{E7E6E6}\textbf{-}     \\ \hline
\end{tabular}}
\label{tab: sytle_loss2}
\vspace{-10pt}
\end{table}

\begin{figure*}[t]
	\centering
	\includegraphics[width=7.1in]
	{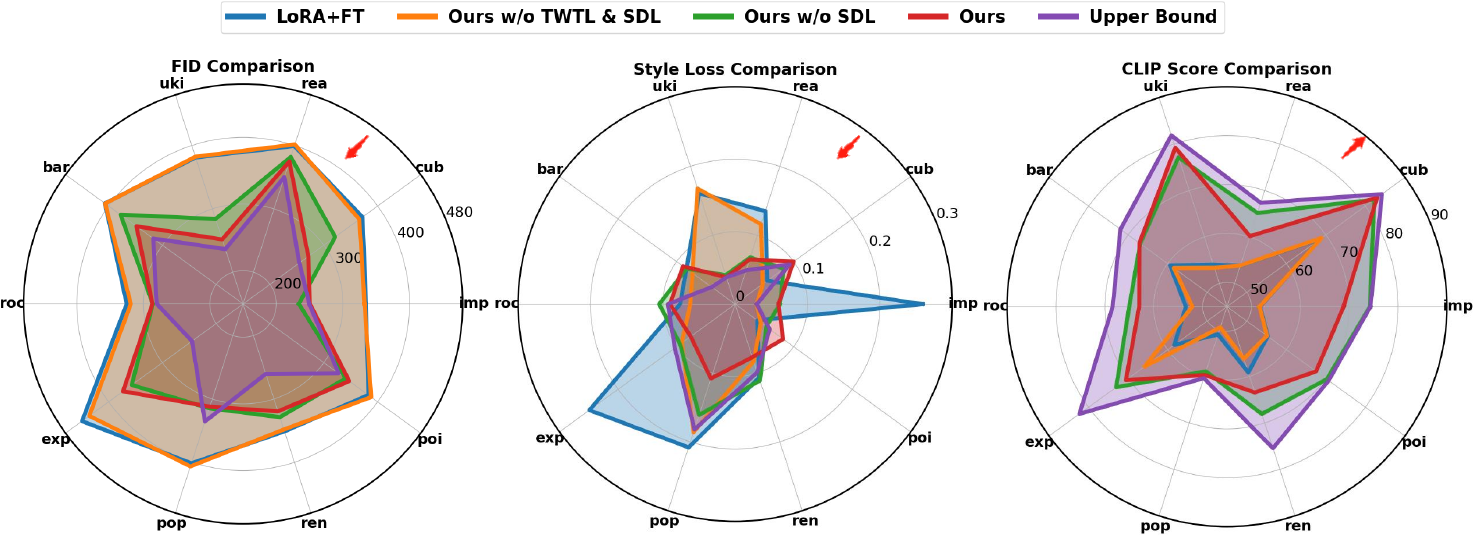}
 	\vspace{-5pt}
	\caption{Style loss, FID and CLIP score for continual style adaptation setting. We conduct ablation studies to evaluate our proposed module one by one. The red arrow in the image indicates the direction where the score improves. Our MuseumMaker achieves the best performance in terms of style loss and FID.} 
	\label{fig: table2}
\end{figure*}

\subsection{Comparison Experiments}
As show in Table~\ref{tab:tab1} and Fig.~\ref{fig: table1}, we present a comprehensive comparison of experiments conducted between our method and other competing approaches. To assess the efficacy of the continual style customization method proposed in this study, we first continually learn all 10 styles available in the Wikiart dataset. Subsequently, we demonstrate the results by generating each style using the same prompt. In order to comprehensively evaluate the performance of the generation, we consider the results of continual style customization from two aspects: qualitative evaluation and quantitative evaluation.

For qualitative evaluation, the results of experiments are shown in Fig.~\ref{fig: experiment1}. The most competitive methods often suffer from catastrophic forgetting of previously encountered styles. For instance, approaches like LoRA combined with fine-tuning (LoRA+FT) and SPD combined with Fine-tuning (SPD+FT) struggle to preserve the ability to generate the textures and color schemes of previous styles. Simply applying the fine-tuning method with LoRA and SPD restricts the T2I diffusion model from retaining prior knowledge. We combine classical continual learning methods EWC and LWF with existing personalized T2I diffusion model to further compare with our method. The combinations such as LoRA+EWC, LoRA+LwF, SPD+EWC, and SPD+LwF vary degrees of forgetting regarding previously learned styles. For instance, the generated results of the pop art style and expressionism style markedly differ from the samples in the dataset. Moreover, the lack of distinct token embeddings for each style in these methods frequently results in confusion among similar styles. Specifically, combinations such as SPD+EWC and SPD+LwF tend to misclassify the baroque style as rococo style, while our method successfully generates accurate results. These outcomes successfully demonstrate that our method adeptly preserves prior knowledge, enabling the generation of a diverse range of styles while effectively distinguishing between similar styles. Our approach showcases a robust capacity to mitigate catastrophic forgetting and differentiate similar stylistic features. This demonstrates the effectiveness of our dual regularization approach, which operates across both the shared-LoRA module and the task-wise token learning module.

For quantitative evaluation, we conduct three metrics (\emph{i.e.,} style loss, FID, CLIP score) to measure the quality of generated images. 
As the result shown in Table.~\ref{tab:tab1}, Table.~\ref{tab: sytle_loss} and Fig.~\ref{fig: table1}, we compare the performance of all the competing methods in terms of style loss, FID and CLIP score, respectively. Our method achieves the best performance, exhibiting improvements ranging from $20.7$ to $67.2$ in terms of FID and from $0.96$ to $11.77$ in terms of CLIP score.  Such improvements demonstrate that our method reduce the catastrophic forgetting to old styles and catastrophic overfitting to new styles. In the realm of style loss, SPD+FT attains the highest performance, as it is directly tailored to optimize style loss. However, when it comes to FID and CLIP scores, it fails to demonstrate satisfactory performance. Our approach ranks second in terms of style loss performance. Compared to LoRA-based methods, SPD+LWF and SPD+EWC, our approach exhibits an improvement ranging from $0.002$ to $0.058$ in style loss. According to the above, our method attains the strongest ability to overcome the catastrophic forgetting and overfitting under the setting of continual style learning for text-to-image diffusion model.

\begin{figure*}[t]
	\centering
	\includegraphics[width=7.1in]
	{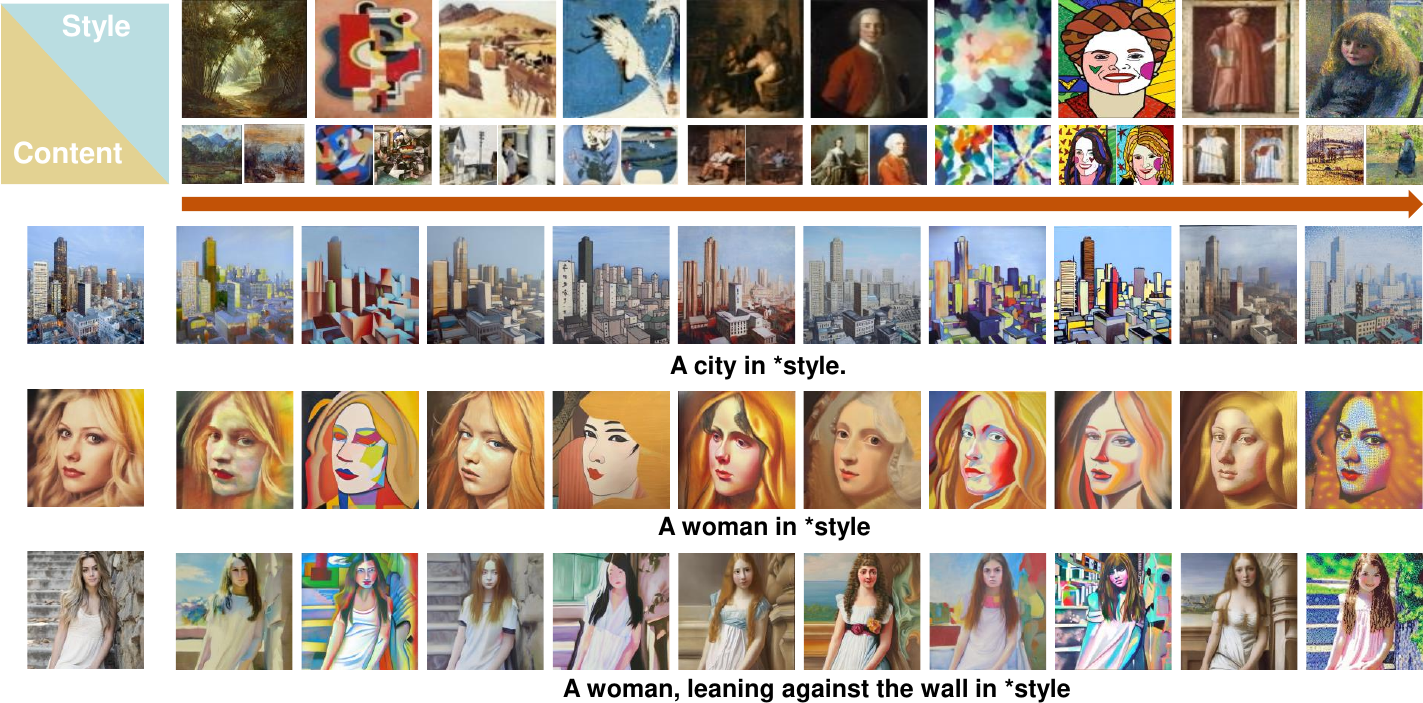}
 	\vspace{-5pt}
	\caption{We conduct the style transfer study with our proposed MuseumMaker, which demonstrates that our method successfully captures various features of 10 styles. } 
	\label{fig: transfer}
\end{figure*}

\subsection{Ablation Studies}
In the subsection, we conduct ablation studies on the continual learning of 10 styles, which encompasses five experimental settings: LoRA+FT, Ours w/o TTL \& SDL, Ours w/o SDL, our complete proposed method and upper bound. For the upper bound setting, we store a unique token embedding and a special LoRA weight for each style. To ensure a fair comparison, all five settings employ the same hyperparameters and number of training steps. Furthermore, we define the same setup as in the comparison experiments to minimize the influence of randomness. Consistent with the comparison experiments, we perform both qualitative and quantitative analyses.

\begin{figure}[t]
	\centering
	\includegraphics[width=3.5in]
	{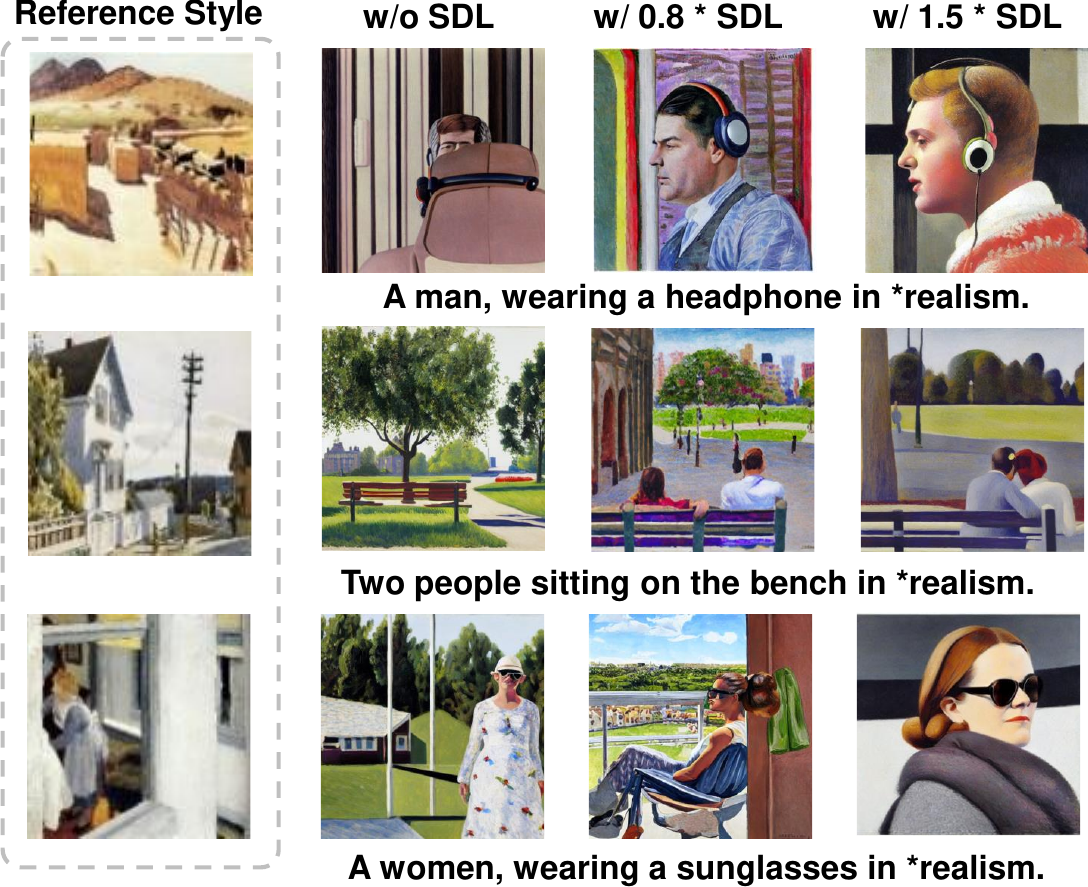}
 	\vspace{-5pt}
	\caption{We compare the image generation performance with varying weights of the SDL module on the realism dataset.} 
	\label{fig: SDL}
\end{figure}

For the qualitative comparison, we evaluate the contribution of each module we proposed, which is depicted in Fig~\ref{fig: experiment2}. We propose a dual regularization for shared-LoRA (DR-LoRA) module and a task-wise token learning (TTL) module to tackle the catastrophic forgetting. We eliminate these two modules one by one to evaluate their effect, and compare the result generated by ours w/o SDL, ours w/o TTL \& SDL and LoRA based fine-tuning (LoRA+FT). As the first three rows of the output section in Fig.~\ref{fig: experiment2}, the images generated by ours w/o SDL shows the best performance on images generation of past styles, which demonstrates the effectiveness of DR-LoRA and TTL module. Additionally, we devise a style distillation loss for addressing catastrophic overfitting. As shown in the last three rows of Fig.~\ref{fig: experiment2}, the lack of the SDL module results in overfitting to the image content compared to our proposed method, adversely affecting the generation of pre-trained concepts. The results generated by our MuseumMaker exhibit a similar performance to the upper bound. To further measure the influence of SDL module under different weights, we design a experiments with varying weights, as shown in Fig.~\ref{fig: SDL}. When the diffusion model has a higher weight of SDL module, the model more effectively reduces the risk of overfitting the content of images. 

\begin{table}[t]
\centering
\setlength{\tabcolsep}{1.0mm}
\renewcommand{\arraystretch}{1.5}
\caption{The comparison of training params and traing time between our method and other competitive method.}
\scalebox{0.92}{
\begin{tabular}{l|>{\centering\arraybackslash}p{3.7cm}>{\centering\arraybackslash}p{3.7cm}}
\hline
Methods & \# Params & Training Time \\ \hline
LoRA+FT            & 39.06M    & 105min        \\
LoRA+LWF           & 39.06M    & 145min        \\
LoRA+EWC           & 39.06M    & 140min        \\
SPD+FT             & 859.52M   & 170min        \\
SPD+LWF            & 859.52M   & 290min        \\
SPD+EWC            & 859.52M   & 179min        \\ \hline
\textbf{Ours}               & 39.06M    & 155min        \\ \hline
Upper Bound        & 390.06M   & 105min        \\ \hline
\end{tabular}}
\label{tab: params}
\vspace{-10pt}
\end{table}

Regarding the quantitative comparison for ablation studies, we evaluate hundreds of generated images with  style loss, FID and CLIP score metrics. As shown in Table.~\ref{tab:tab2},Table.~\ref{tab: sytle_loss2}  and Fig.~\ref{fig: experiment2}, we compare ours, ours w/o SDL, our w/o TTL \& SDL and LoRA+FT with three metrics to evaluate the effectiveness of each proposed module. Ours method achieves the best performance on two metrics (\emph{i.e.,} style loss and FID), which improve $0.007\sim0.058$ and $9.0\sim66.9$ in terms of styles loss and FID, respectively. As we propose DR-LoRA and TTL to tackle catastrophic forgetting, the setting of ours w/o SDL and our w/o TTL \& SDL increase $0.015\sim0.051$ in terms of style loss, $1.8\sim57.9$ in terms of FID and $0.01\sim13.61$ in terms of CLIP score. The improvement across these three metrics highlights the effectiveness of the DR-LoRA module and TTL module in addressing catastrophic forgetting. However, the setting of ours w/o SDL achieves the best performance on CLIP score, and our method falls short by 1.84. Lower scores of ours w/o SDL in terms of style loss and FID reveal that ours w/o SDL slightly overfits to the content of images and learns impure represents of styles, which leads to a better adaptability to CLIP score. This phenomenon justifies the importance of SDL module to overcome catastrophic overfitting.


\subsection{Model Efficiency Study}
To thoroughly evaluate the efficiency and effectiveness of our proposed MuseumMaker against other competitive methods, we detailedly analyze the training parameters and training time of comparison methods, as presented in Table.\ref{tab: params}. While LoRA-based methods require a substantial number of parameters (39.06M) and training times ranging from 105 to 145 minutes, these methods show unsatisfactory generation results for continual style customization. Moreover, SPD-based methods utilize a much larger number of parameters (859.52M) and a relatively long training time, ranging from 170 to 290 minutes. Due to such a large amount of computing consumption, SPD-based methods are not efficient enough for continual style customization task. Our approach achieves comparable performance with significantly minimal parameters (39.06M) and a relatively short training time of 155 minutes. Considering the results of image generation, our approach stands out for achieving performance closest to the upper bound with minimal training parameters. These outcomes demonstrate the remarkable effectiveness of our DR-LoRA, task-wise token learning and style distillation loss techniques in the context of continuous style customization tasks, showcasing its potential for practical applications in real-world scenarios.

\subsection{Style Transfer Study}
To evaluate the efficacy of our proposed continuous style customization approach in capturing the intricate nuances and diverse characteristics of styles continually provided by users, we devise its application in style transfer studies. This expansion not only broadens the scope of our method but also demonstrates its adaptability to diverse tasks in the field of computer vision. The exploration leverages the accumulated knowledge captured from the continuous stream of style data to facilitate the transfer of style from input images. As depicted in Fig.\ref{fig: transfer},  we observe that MuseumMaker adeptly captures the subtle textures, color themes, and structural elements after obtaining knowledge from 10 distinct styles. These results demonstrate the remarkable effectiveness of MuseumMaker in seamlessly learning a multitude of styles for both image generation and image style transfer, showcasing its versatility beyond the realm of text-to-image tasks.

\section{Conclusion}
This paper explores the challenge of continually learning new user-provided styles and generating a variety of stylistic images with a pre-trained diffusion model. Our proposed MuseumMaker efficiently embeds styles into diffusion model without overfitting to the content of images. Specially, we consider two critical aspects in developing our MuseumMaker, \emph{i.e.,} ``catastrophic forgetting'' and ``catastrophic overfitting''. We devise a dual regularization for shared-LoRA module and a task-wise token learning module to address the ``catasrtophic forgetting''. These modules help the model retain knowledge of previously encountered styles while adapting to new ones. Furthermore, we adopt a style distillation loss to tackle the ``catastrophic overfitting''. This loss function ensures that the model learns pure representation of styles without being overly influenced by the content of images. Comprehensive experiments are conducted to show the effectiveness of our method. Results show that our method outperforms existing approaches in terms of style loss, FID and CLIP score, highlighting its ability to generate diverse and high-quality stylistic images.



\bibliographystyle{IEEEtran}
\bibliography{ref}{}

\newpage

\end{document}